\documentclass[10pt,twocolumn,letterpaper]{article}

\usepackage{cvpr}    

\usepackage{xcolor}

\usepackage[ruled,vlined,linesnumbered]{algorithm2e}
\usepackage{amsmath, amssymb, amsfonts, amsthm, mathrsfs}
\DeclareMathOperator*{\argmax}{arg\,max}
\usepackage{multirow, multicol, booktabs}
\usepackage[table]{xcolor}
\usepackage{placeins}
\usepackage{float}
\usepackage{subcaption}
\usepackage{xcolor}

\definecolor{ourlightcyan}{RGB}{210, 245, 255}
\definecolor{ourlightred}{RGB}{255, 220, 220}
\definecolor{ForestGreen}{RGB}{34,139,34}
\definecolor{cvprblue}{rgb}{0.21,0.49,0.74}
\colorlet{TableColor}{ourlightcyan}
\colorlet{MissingColor}{ourlightred}
\definecolor{mygreen}{RGB}{0,246,0}
\definecolor{myorange}{RGB}{255,102,20}

\newcommand{\green}[1]{\textcolor{mygreen}{#1}}
\newcommand{\orange}[1]{\textcolor{myorange}{#1}}

\newcommand{\std}[2]{#1 $\pm$ #2}
\DeclareMathOperator{\expmap}{expm}

\usepackage[pagebackref,breaklinks,colorlinks,allcolors=cvprblue]{hyperref}

\def\paperID{XXXX}   
\def\confName{CVPR}
\def\confYear{2026}

\title{A Hyperbolic Perspective on Hierarchical Structure in Object-Centric Scene Representations}

\vspace{-0.6cm}

\author{
Neelu Madan$^{1}$\thanks{Corresponding author: \href{mailto:neeluk.madan@gmail.com}{neeluk.madan@gmail.com}} \quad
Àlex Pujol Vidal$^{1}$ \quad
Andreas Møgelmose$^{1}$ \quad
Sergio Escalera$^{1,2}$ \\
Kamal Nasrollahi$^{1,3}$ \quad
Graham W. Taylor$^{4}$ \quad
Thomas B. Moeslund$^{1}$
\\[0.5em]
{\small $^1$Aalborg University \& Pioneer Centre for AI, Denmark \quad
$^2$Universitat de Barcelona \& Computer Vision Center, Spain } \\
{\small $^3$Milestone Systems, Denmark \quad
$^4$University of Guelph \& Vector Institute, Canada 
\vspace*{-0.3cm}
}
}

\begin{document}
\maketitle

\vspace{-0.5cm}

\begin{abstract}
Slot attention has emerged as a powerful framework for unsupervised object-centric learning, decomposing visual scenes into a small set of compact vector representations called \emph{slots}, each capturing a distinct region or object. However, these slots are learned in Euclidean space, which provides no geometric inductive bias for the hierarchical relationships that naturally structure visual scenes. In this work, we propose a simple post-hoc pipeline to project Euclidean slot embeddings onto the Lorentz hyperboloid of hyperbolic space, without modifying the underlying training pipeline. We construct five-level visual hierarchies directly from slot attention masks and analyse whether hyperbolic geometry reveals latent hierarchical structure that remains invisible in Euclidean space. Integrating our pipeline with SPOT~\cite{kakogeorgiou2024spot} (images), VideoSAUR~\cite{Zadaianchuk-NeurIPS-2023} (video), and SlotContrast~\cite{Manasyan-CVPR-2025} (video), We find that hyperbolic projection exposes a consistent scene-level to object-level organisation, where coarse slots occupy greater manifold depth than fine slots, which is absent in Euclidean space. We further identify a "curvature--task tradeoff": low curvature ($c{=}0.2$) matches or outperforms Euclidean on parent slot retrieval, while moderate curvature ($c{=}0.5$) achieves better inter-level separation. Together, these findings suggest that slot representations already encode latent hierarchy that hyperbolic geometry reveals, motivating end-to-end hyperbolic training as a natural next step. Code and models are available at \href{https://github.com/NeeluMadan/HHS}{github.com/NeeluMadan/HHS}.
\vspace*{-0.3cm}
\end{abstract}

\section{Introduction}


Understanding visual scenes requires decomposing them into meaningful, structured components such as objects, parts, and reasoning about their relationships.
Learning such representations is studied in the literature under the umbrella of object-centric learning (OCL)~\cite{Locatello-NeurIPS-2020, Greff-ICML-2019, Greff-NeurIPS-2016, sabour2017dynamic}, 
spanning amortized grouping~\cite{Greff-NeurIPS-2016}, 
generative approaches~\cite{Burgess-arxiv-2019,Greff-ICML-2019}, 
routing-based methods~\cite{sabour2017dynamic, hinton2018matrix, kosiorek2019stacked}, 
and slot attention~\cite{Locatello-NeurIPS-2020, Seitzer-ICLR-2022, Singh-NeurIPS-2022, Elsayed-NeurIPS-2022}.
Among these, slot attention has emerged as the most successful paradigm, representing each scene element as a compact ``slot'' vector learned via iterative attention over visual features.
Recent methods further combine slot attention with powerful pretrained visual features~\cite{Seitzer-ICLR-2022, kakogeorgiou2024spot, Zadaianchuk-NeurIPS-2023}, achieving strong unsupervised segmentation performance on complex real-world images and videos.

\begin{figure}[!t]
    \centering
    \includegraphics[width=0.5\textwidth]{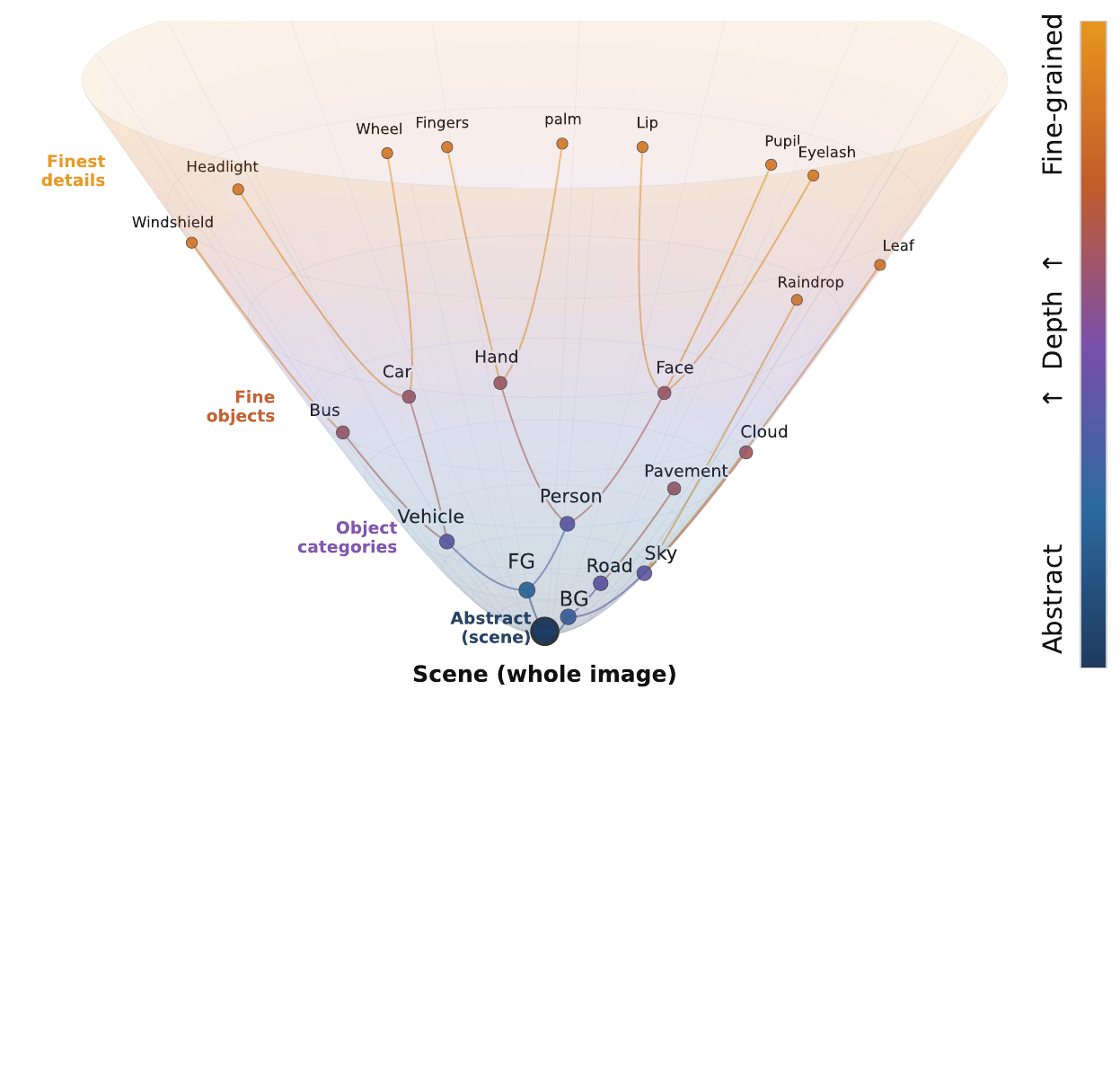} 
    \vspace{-3.5cm}
    \caption{
        Visual hierarchy on the Lorentz hyperboloid. Abstract scene-level concepts reside near the apex; representations grow increasingly fine-grained with depth, with geodesics encoding parent--child relationships.
        This geometric property motivates our analysis of object-centric slot representations through a hyperbolic lens.
    }
    \label{fig:teaser}
    \vspace{-0.4cm}
\end{figure}

Despite this progress, a fundamental limitation remains. Slots are represented as points in Euclidean space, which treats all scene elements as geometrically equivalent. Yet visual scenes are inherently hierarchical: a whole object contains parts, and coarse categories contain fine-grained instances. Euclidean geometry lacks the inductive bias to capture these relationships, leaving hierarchical structure implicit and unrepresented.
Hyperbolic spaces, characterized by constant negative curvature and exponential volume growth with radius, are natural containers for such hierarchical data. Originally demonstrated in NLP~\cite{nickel2017poincare}, the benefits of hyperbolic geometry have since extended to graphs~\cite{liu2019hyperbolic}, images~\cite{atigh2022hyperbolic}, and open-vocabulary recognition~\cite{desai2023hyperbolic}, consistently showing advantages when latent structure is hierarchical.
Motivated by this, we ask: \textit{\textit{do object-centric slot representations, learned in Euclidean space, carry latent hierarchical structure (as illustrated in Figure \ref{fig:teaser}) that is revealed through a hyperbolic lens?}}

In this work, we propose a post-hoc pipeline to project Euclidean slot representations into the hyperbolic space and study whether hyperbolic geometry better reveals the hierarchical structure latent in object-centric representations.
We construct a five-level visual hierarchy from slot attention masks at different granularities and measure whether hyperbolic distances between parent and child slots are more faithful to this visual hierarchy than their Euclidean counterparts.
Because we do \emph{not} modify the Euclidean training pipeline, our approach applies to any slot attention framework.
We validate this on established baselines: SPOT~\cite{kakogeorgiou2024spot} (image), VideoSAUR~\cite{Zadaianchuk-NeurIPS-2023} (video), and SlotContrast \cite{Manasyan-CVPR-2025} (video). Projecting onto the Lorentz hyperboloid exposes a consistent coarse-to-fine organization among slots that is hardly visible in the original Euclidean space


\noindent\textbf{Contributions:}
\begin{itemize}
\item We demonstrate that slot representations encode latent hierarchy, accessible through hyperbolic projection barely visible in their native Euclidean space, opening a previously unexplored axis of analysis for object-centric learning.
\item We find a consistent inverted depth ordering across two datasets and three models: coarse slots occupy greater manifold depth than fine slots, reflecting the larger representational volume that scene-level context requires.
\item We identify a \emph{curvature--task tradeoff}: low curvature ($c{=}0.2$) matches or outperforms Euclidean on parent retrieval, while increasing curvature ($c{=}0.5$ or $c{=}1.0$) achieves better level separation, providing practical guidelines for future end-to-end hyperbolic OCL.
\end{itemize}

\section{Related Work}

\noindent
\textbf{Object-Centric Learning.}
Object-centric learning aims to decompose scenes into constituent objects and their properties without direct supervision. Early approaches such as MONet~\cite{Burgess-arxiv-2019}, IODINE~\cite{Greff-ICML-2019}, and GENESIS~\cite{engelcke2020genesis} employed Variational Autoencoders~\cite{kingma2014auto} to model scenes as mixtures of component distributions.
More recent work shifted toward attention-based approaches, with Slot Attention~\cite{Locatello-NeurIPS-2020} introducing competitive iterative attention for object binding. This paradigm has been extended in many directions: 
to real-world scenes by using self-supervised DINO features 
as reconstruction targets~\cite{Seitzer-ICLR-2022}, 
to video understanding~\cite{kipf-ICLR-2022,Elsayed-NeurIPS-2022,Zadaianchuk-NeurIPS-2023,Aydemir-NeurIPS-2023,Manasyan-CVPR-2025,zhao2025slot}, 
to autoregressive generation~\cite{Elsayed-NeurIPS-2022}, 
to improved spatial binding~\cite{kakogeorgiou2024spot}, 
to learnable slot initialization~\cite{Jia-ICLR-2023}, 
to self-supervised representation learning~\cite{singh2022illiterate}, 
to general-purpose object-centric representations~\cite{Didolkar2024ZeroShotOCRL,zhao2025vector}, 
and to 3D scenes~\cite{liu2024slotlifter,smith2023unsupervised,yu2022unsupervised,sajjadi2022object}.
Despite these advances, all of the above represent slots as fixed-dimensional vectors in Euclidean space, providing no geometric inductive bias toward the hierarchical structure inherent in natural scenes, a gap our work addresses. 
Our work is also related to video object segmentation benchmarks for complex scene understanding, including MOSE~\cite{ding2023mose}, MOSEv2~\cite{ding2025mosev2}, MeViS~\cite{ding2023mevis}, and MeViSv2~\cite{ding2025mevis}. Unlike these supervised benchmarks, our work is entirely unsupervised and studies latent hierarchical structure in slot representations rather than segmentation performance.

\noindent
\textbf{Hyperbolic Representation Learning.} Hyperbolic spaces are Riemannian manifolds of constant negative curvature whose volume grows \emph{exponentially} with radius, making them a natural fit for embedding hierarchical and tree-like structures that would require exponentially large Euclidean spaces~\cite{nickel2017poincare}.
Poincaré Embeddings~\cite{nickel2017poincare} first demonstrated that hyperbolic representations outperform Euclidean ones for hierarchical NLP data such as WordNet.
Hyperbolic Neural Networks~\cite{ganea2018hyperbolic} extended this to deep learning by defining gyrovector-space analogues of fully-connected layers, activations, and attention operations 
entirely within the Poincaré ball.
More recently, Fully Hyperbolic Neural Networks~\cite{chen2022fully} reformulated convolutional and linear layers using the Lorentz model, enabling end-to-end hyperbolic training for vision tasks.
In the vision domain, hyperbolic representations have been applied to 
few-shot learning~\cite{gao2021curvature, li2025hypdae}, 
image segmentation~\cite{atigh2022hyperbolic, sur2025hyperbolic}, 
and open-vocabulary recognition~\cite{desai2023hyperbolic, poppi2025hyperbolic}, 
consistently showing benefits when the label space or feature space has a latent hierarchical structure.

\noindent
\textbf{Hierarchical Structure in Visual Scenes.}
The compositional and part-whole nature of visual scenes has long motivated hierarchical approaches in computer vision, from part-based models~\cite{hinton2023represent} 
and grouping and attention aggregation transformers~\cite{xu2022groupvit,aasan2025differentiable} 
to capsule networks~\cite{sabour2017dynamic,kosiorek2019stacked,hinton2018matrix}.
More recently, hyperbolic geometry~\cite{wang2025learning, qiu2024hihpq, wang2023hi}  has been explored as a principled space to encode such visual hierarchies. 
To the best of our knowledge, no prior work has studied hierarchical structure within object-centric slot representations in either Euclidean or hyperbolic space.
Our work is the first to ask whether the latent hierarchy among slots is better revealed through a hyperbolic lens than in their native Euclidean embedding space.

\begin{figure*}[t]
    \centering
    \vspace{-0.2cm}
    \includegraphics[width=0.9\textwidth]{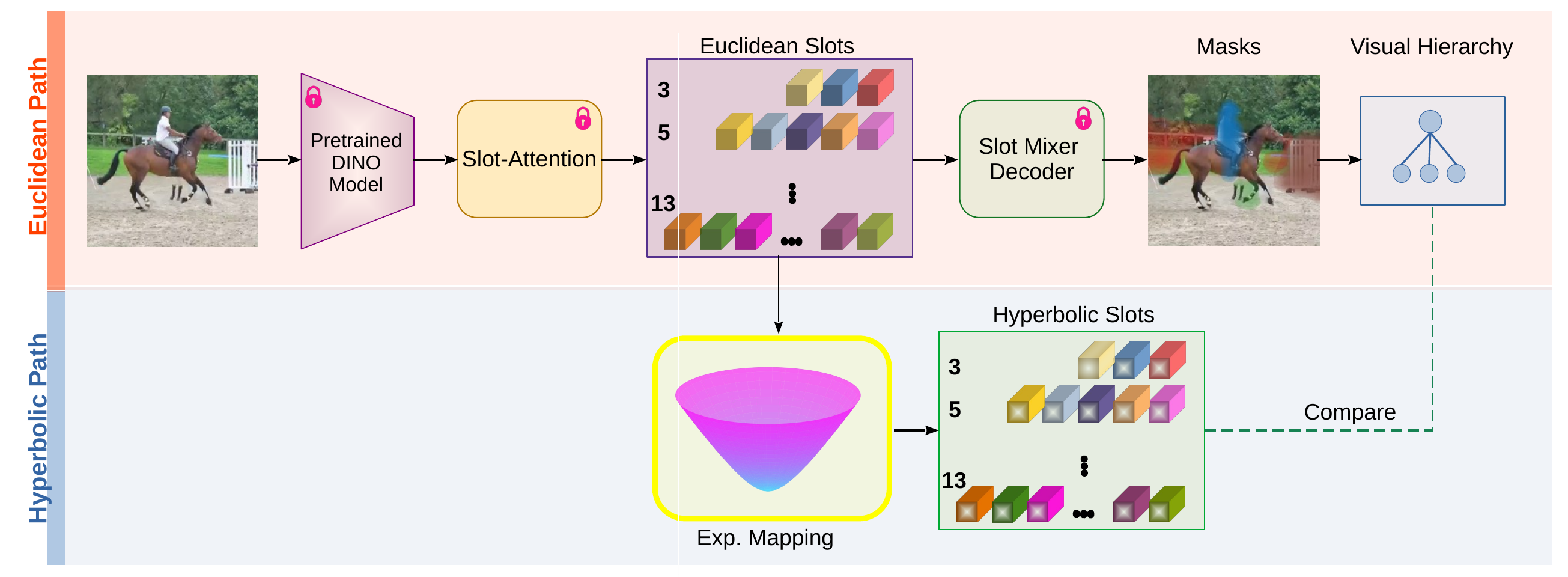} 
    \vspace{-0.2cm}
    \caption{
    Overview of our post-hoc pipeline.
    Patch features from a frozen DINOv2 backbone are decoded into
    $N$ slot representations via Slot Attention.
    The Euclidean path reconstructs slot masks through the baseline decoder, from which ground-truth parent--child pairs $\mathcal{P}$ are derived via mask inclusion.
    The hyperbolic path projects the same slots onto the Lorentz hyperboloid $\mathbb{H}^d_K$ via the exponential map, where geodesic distances and Lorentz norms are used for hierarchical analysis without any modification to the original training.
    }
    \label{fig:pipeline}
    \vspace{-0.4cm}
\end{figure*}

\section{Method}

Our goal is to investigate whether hyperbolic geometry reveals hierarchical structure in object-centric slot representations more clearly than Euclidean geometry.
Rather than modifying the training pipeline, we obtain slots from an existing Euclidean framework, project them into hyperbolic space, and measure whether parent--child relationships between slots at different levels of decomposition are better captured by hyperbolic distances than Euclidean ones.
Figure~\ref{fig:pipeline} illustrates the overall approach.

\subsection{Feature Extraction and OCL Framework.}
\label{sec:slots}

We extract patch-level features from input images (or video frames) using a frozen pretrained DINOv2 backbone~\cite{caron2021emerging}. For an image of resolution $H \times W$ with patch size $p$, we obtain feature tokens $\mathbf{F} \in \mathbb{R}^{L \times d_f}$ where $L = HW/p^2$, and $d_f$ is the dimension of DINO features.

Following~\cite{Locatello-NeurIPS-2020}, we apply slot attention with $N$ randomly initialized slot vectors $\{\mathbf{s}_i\}_{i=1}^N \subset \mathbb{R}^{d_s}$ as queries over $\mathbf{F}$. Slots compete via a softmax normalization over the slot dimension, encouraging each slot to specialise to a distinct scene region. After $T$ attention iterations, we obtain slot representations $\mathbf{S} = [\mathbf{s}_1, \ldots, \mathbf{s}_N] \in \mathbb{R}^{N \times d_s}$, along with per-slot attention masks $\mathbf{M} \in [0,1]^{N \times L}$ that spatially localise each slot's region of influence.

All baseline models are trained end-to-end via a reconstruction loss on DINOv2 features, using the decoder of each respective model: an MLP decoder for VideoSAUR~\cite{Zadaianchuk-NeurIPS-2023} and SlotContrast~\cite{Manasyan-CVPR-2025}, and an autoregressive decoder for SPOT~\cite{kakogeorgiou2024spot}. In our analysis, all model components, including backbone, slot attention module, and decoder, are kept entirely frozen. We use the pretrained models solely to extract slot representations $\mathbf{S}$ and attention masks $\mathbf{M}$ at inference time.

\subsection{Building the Visual Hierarchy}
\label{sec:hierarchy}

To study hierarchical structure, we decompose each scene at \emph{five granularities} $N \in \{3, 5, 7, 11, 13\}$, yielding a sequence of slot sets $\mathbf{S}^{(N)}$ with corresponding masks $\mathbf{M}^{(N)}$. We evaluate all consecutive coarse--fine pairs $(N_1, N_2) \in \{(3,5),\, (5,7),\, (7,11),\, (11,13)\}$, where $N_1$ is the coarse level and $N_2$ the fine level. We evaluate consecutive level pairs as the simplest test of local hierarchical structure; non-consecutive pairs (e.g., $N=3\rightarrow7$) and multi-level chains are left for future work. 

\noindent\textbf{Parent--child assignment.}
For each pair $(N_1, N_2)$, we define a \textbf{parent--child} relationship between a coarse slot $\mathbf{s}_i^{(N_1)}$ and a fine slot $\mathbf{s}_j^{(N_2)}$ based on spatial mask inclusion in pixel-space.
We threshold both masks to obtain binary segmentation regions $\hat{\mathbf{m}}_i^{(N_1)}, \hat{\mathbf{m}}_j^{(N_2)} \in \{0,1\}^L$, and compute the inclusion score of each fine slot $j$ with respect to each coarse slot $i$ as:
\vspace{-0.3cm}
\begin{equation}
    \mathcal{I}(j, i) = 
    \frac{\hat{\mathbf{m}}_j^{(N_2)} \cdot \hat{\mathbf{m}}_i^{(N_1)}}
         {\|\hat{\mathbf{m}}_j^{(N_2)}\|_1}
    \label{eq:inclusion}
\end{equation}
A threshold-based assignment ($\mathcal{I}(j,i) \geq \tau$) can leave fine slots unassigned when no coarse slot meets the threshold, producing orphan children that cannot participate in hierarchical evaluation. To avoid this, we assign each fine slot to its best-matching coarse slot via:
\vspace{-0.2cm}
\begin{equation}
    \text{parent}(j) = \argmax_{i} \; \mathcal{I}(j, i)
    \vspace{-0.2cm}
    \label{eq:assignment}
\end{equation}
This guarantees every fine slot has exactly one parent, yielding a complete set of parent--child pairs $\mathcal{P}^{(N_1, N_2)} = \{(\text{parent}(j),\, j)\}_{j=1}^{N_2}$ for each consecutive level pair. Fine slots whose inclusion score with their assigned parent exceeds $\tau_{\text{excl}} = 0.95$ are considered near-identical to their parent. This indicates that the slot has not split into a more fine-grained region. Such slots are therefore excluded from evaluation.
We restrict evaluation to consecutive level pairs $(N_c, N_f)$ where $N_f$ is the next granularity above $N_c$, as these transitions present the hardest retrieval setting: the structural gap between adjacent levels is small, making distance-based discrimination more challenging than for non-consecutive pairs such as $(3, 13)$ where the representational difference is larger. Evaluating on non-consecutive pairs and full leaf-to-root chains is left as future work.

\subsection{Projection to Hyperbolic Space}
\label{sec:hyperbolic}

We project each slot $\mathbf{s}_i \in \mathbb{R}^{d_s}$ to the Lorentz hyperboloid $\mathcal{L}^{d_s}_c = \{\mathbf{x} \in \mathbb{R}^{1+d_s} : \langle \mathbf{x}, \mathbf{x} \rangle_{\mathcal{L}} = -1/c\}$, with $c > 0$ and $\langle . , .\rangle_\mathcal{L}$ the Lorentzian inner product, using the exponential map at the origin $\mathbf{o} = (1/\sqrt{c}, \mathbf{0})\in \mathbb{R}^{1+d_s}$:
\vspace{-0.3cm}
\begin{equation}
    \begin{aligned}
    \mathbf{s}_i^{(\mathcal{L})} = \expmap^c_{\mathbf{o}}(\mathbf{s}_i) 
    &= \cosh\!\left(\sqrt{c}\,\|\mathbf{s}_i\|\right)\mathbf{o} \\
    &\quad + \sinh\!\left(\sqrt{c}\,\|\mathbf{s}_i\|\right)
      \frac{\mathbf{s}_i}{\sqrt{c}\|\mathbf{s}_i\|},
    \end{aligned}
    \label{eq:expmap}
    \vspace{-0.2cm}
\end{equation}
where $c \in \{0.2, 0.5, 1\}$ is the fixed curvature of the hyperboloid, and $\|\cdot\|$ denotes the Euclidean norm. Note that $\mathbf{s}_i^{(\mathcal{L})} \in \mathcal{L}^{d_s}_c \subset \mathbb{R}^{1+d_s}$.

\subsection{Distance Metric}
\label{sec:distance}

We use three geometric measures across our analyses: cosine distance for parent retrieval, and Euclidean and Lorentz norms for level separation.

\noindent\textbf{Cosine distance.} Given two slot representations $\mathbf{s}_i, \mathbf{s}_j \in \mathbb{R}^d$, we measure their similarity in Euclidean space via cosine distance. Slots are first $\ell_2$-normalized, and the distance is defined as:
\vspace{-0.4cm}
\begin{equation}
    d_{\cos}(\mathbf{s}_i, \mathbf{s}_j) = 1 - \frac{\mathbf{s}_i^\top \mathbf{s}_j}{\|\mathbf{s}_i\| \|\mathbf{s}_j\|}.\
\end{equation}

This metric captures angular similarity between slot vectors and is used for parent slot retrieval (Section~\ref{sec:retrieval}), motivated by CutLER~\cite{wang2023cut} which measures patch-wise feature similarity at the slot level.

\noindent\textbf{Euclidean norm.} We also analyse the raw $\ell_2$ norm $\|\mathbf{s}_i\|_2$ of slot vectors before normalization, as a baseline signal for level separation. Since training never supervises slot norms directly, this quantity carries no guaranteed geometric meaning; we include it to examine whether any incidental norm-based structure emerges across granularity levels.

\noindent\textbf{Lorentz distance.} We take the same two slot representations and project them to $\mathcal{L}^d_s$ by means of the exponential map (Eq.~\ref{eq:expmap}), and obtain $\mathbf{s}^{(\mathcal{L})}_i, \mathbf{s}^{(\mathcal{L})}_j \in \mathbb{R}^{1+d}$. We aim to capture the hierarchical structure that emerges from the pretrained slots. Thus, we use the Lorentz distance between hyperbolic embeddings of the slots, defined as:
\vspace{-0.3cm}
\begin{equation}\label{eq:dist_lor}
    d_{\mathcal{L}_c^d}(\mathbf{s}^{(\mathcal{L})}_i, \mathbf{s}^{(\mathcal{L})}_j) = \frac{1}{\sqrt{c}} \cosh^{-1}\!\left(-c \langle \mathbf{s}^{(\mathcal{L})}_i, \mathbf{s}^{(\mathcal{L})}_j \rangle_{\mathcal{L}}\right).
\end{equation}

Unlike cosine distance, $d_\mathcal{L}$ is sensitive to the depth of a point on the manifold and scales with the curvature $c$. 
Such a distance is used for level separation analysis (Section~\ref{sec:dilatation}), while $d_{\mathcal{L}}$ is also evaluated for parent retrieval (Section~\ref{sec:retrieval}).

\subsection{Centroid Computation}

To measure separation between hierarchy levels we aggregate the slots of each level obtaining a level representation vector for each geometry. For the Euclidean case, this is obtained by simply computing the mean for each level. In the hyperbolic case we compute the center of mass of the Lorentzian inner product \cite{pmlr-v97-law19a}, given by:
\vspace{-0.3cm}
\begin{equation}\label{eq:centroid}
    \mathbf{\mu}_k^{(\mathcal{L})} = \frac{\sum_{i_k=1}^{N_k}\mathbf{s}^{(\mathcal{L})}_{i_k}/N_k}{\sqrt{c}|\|\sum_{i_k=1}^{N_k}\mathbf{s}^{(\mathcal{L})}_{i_k}/N_k\|_{\mathcal{L}}|},
    \vspace{-0.2cm}
\end{equation}
where $|\|\cdot\|_{\mathcal{L}}| = \sqrt{\|\cdot\|_{\mathcal{L}}^2}$ is the modulus of the imaginary Lorentzian norm, and $k$ indicates the granularity level of the hierarchy that has $N_k$ slots.

\noindent\textbf{Analysis protocol.}
For each ground-truth level of parent--child pairs $(\mathbf{s}^{(c)}_i, \mathbf{s}^{(f)}_j)$ derived from mask inclusion (Section~\ref{sec:hierarchy}), we compute $d_{\mathcal{L}}$ between the centroids of projected slot level-pairs and analyse whether the Lorentz distance to the origin reveal consistent hierarchical structure across granularity levels, geometries, and baselines. 




\vspace{-0.1cm}
\section{Experiments}
\label{sec:experiments}
\vspace{-0.1cm}

\noindent\textbf{Baselines.}
We extract slot representations from three object-centric learning baselines: VideoSAUR~\cite{Zadaianchuk-NeurIPS-2023} (video) and SlotContrast~\cite{Manasyan-CVPR-2025} (video) on the YTVIS 2021 validation set~\cite{yang2019video}, and SPOT~\cite{kakogeorgiou2024spot} (image) on the MS-COCO validation set~\cite{lin2014microsoft}. All models use a DINOv2~\cite{oquab2023dinov2} backbone; since the original VideoSAUR was trained with DINOv1~\cite{carion2020end} features, we use a VideoSAUR variant retrained with DINOv2 for consistency.
For video models, we use slots from the last frame of each video, as temporal recurrence refines slot assignments across frames and the final frame thus yields the most stable scene decomposition. Analysing deeper hierarchical dynamics along the temporal axis is left as future work.
For all baselines, we use pretrained weights and perform inference only, extracting slot representations at five granularity levels $N \in \{3, 5, 7, 11, 13\}$ without any retraining, by varying the number of slots at inference time. 

\noindent\textbf{Evaluation metric.} Hit@1 measures the percentage of fine slots for which the GT parent is ranked first among all coarse slots by distance. Higher number is better for this metric, with a random baseline of $1/N_k$.

\begin{figure*}[!t]
    \centering
    \begin{minipage}[b]{0.32\textwidth}
        \centering
        \includegraphics[width=\textwidth]{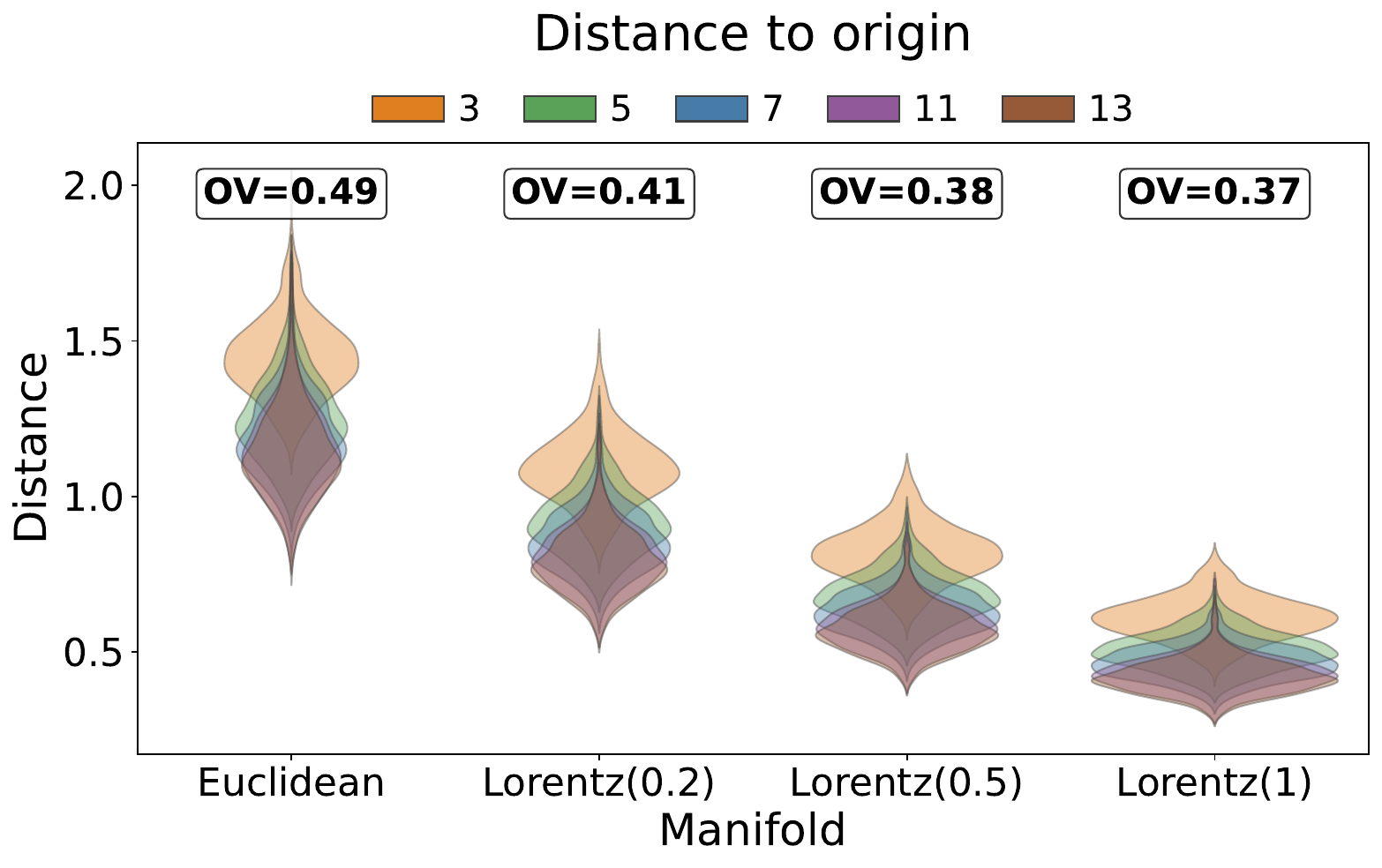}
        \subcaption{SlotContrast (YTVIS)}
        \label{fig:violin_slotcontrast}
    \end{minipage}
    \hfill
    \begin{minipage}[b]{0.32\textwidth}
        \centering
        \includegraphics[width=\textwidth]{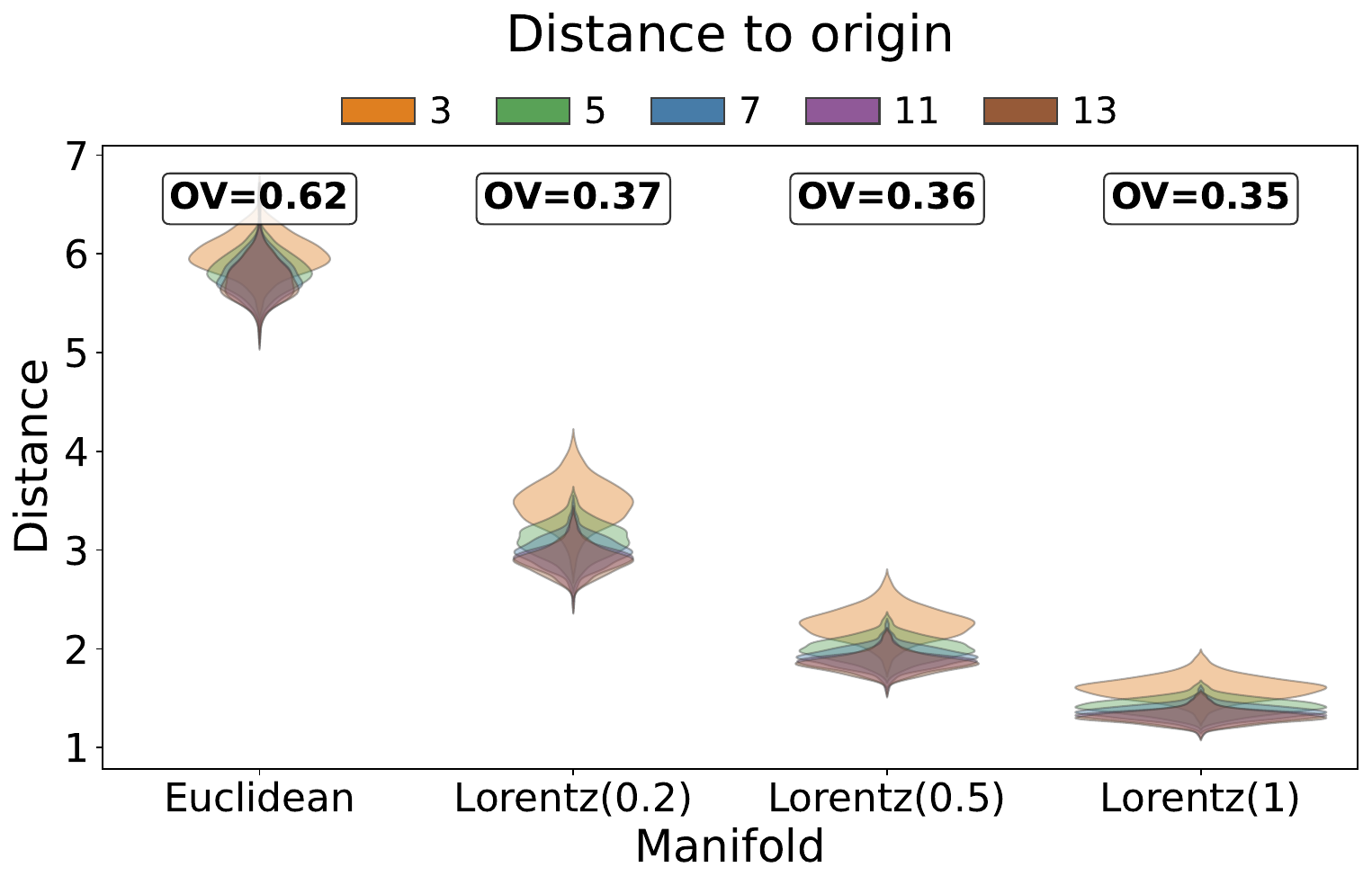}
        \subcaption{VideoSAURv2 (YTVIS)}
        \label{fig:violin_videosaur}
    \end{minipage}
    \hfill
    \begin{minipage}[b]{0.32\textwidth}
        \centering
        \includegraphics[width=\textwidth]{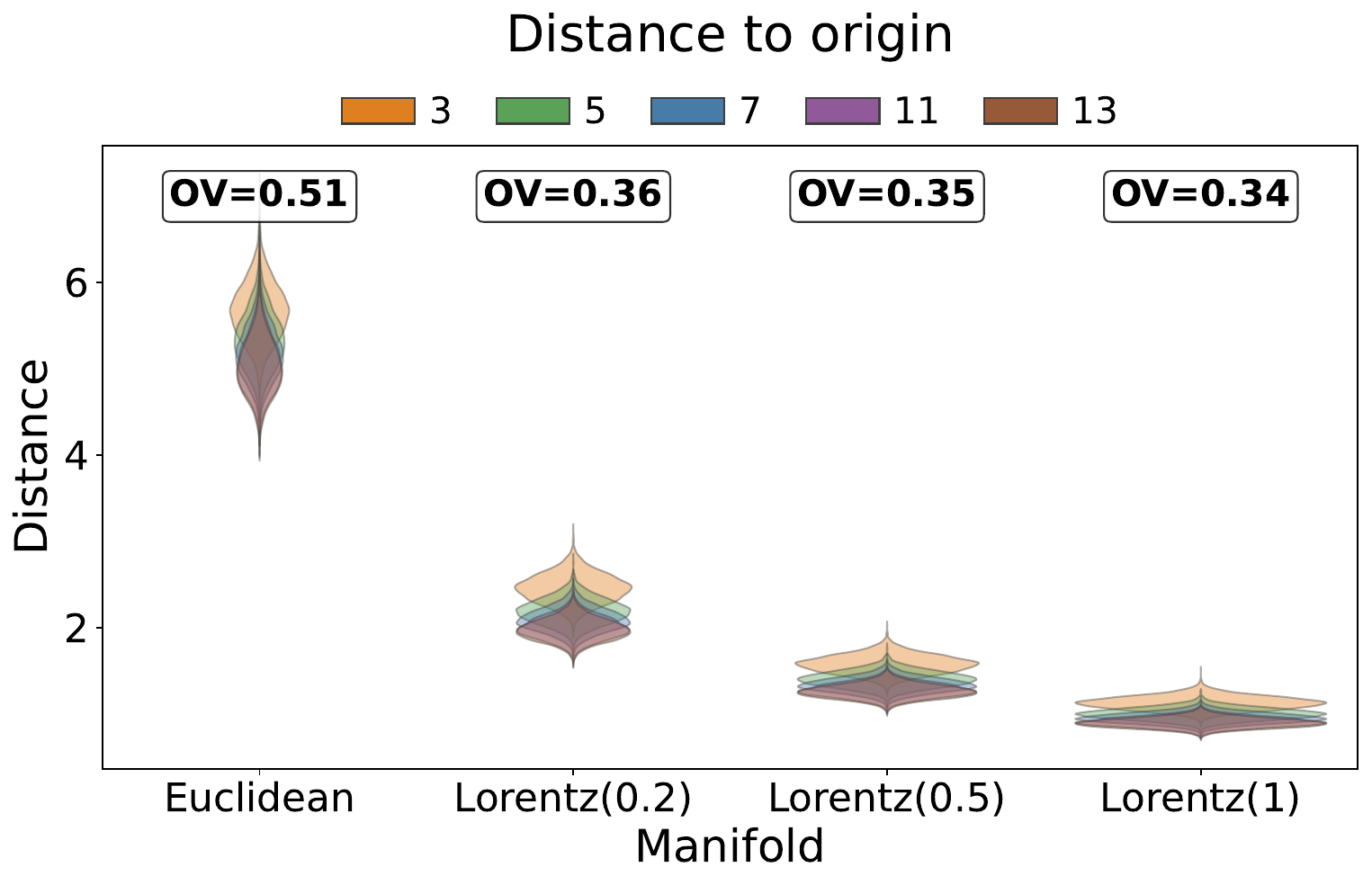}
        \subcaption{SPOT (COCO)}
        \label{fig:violin_spot}
    \end{minipage}
    \caption{
        Distribution of 300 videos frames (\ref{fig:violin_slotcontrast}, \ref{fig:violin_videosaur}) and 5000 images (\ref{fig:violin_spot}) of distances from centroid to origin across levels $N \in \{3,5,7,11,13\}$ for three baselines, and level-wise overlap score OV. Lower OV indicates better level separation. $c{=}0.5$ offers the best separation-to-compression tradeoff; higher curvatures achieve tighter but increasingly compressed distributions. The depth ordering is consistently inverted across all models: coarse slots lie furthest from the origin, fine slots closest.
    }
    \vspace{-0.4cm}
    \label{fig:violin}
\end{figure*}

\subsection{Level Separation in Hyperbolic Space}
\label{sec:dilatation}


\noindent\textbf{Setup.}
For each granularity level $N_k \in \{3, 5, 7, 11, 13\}$, we measure the mean distance to the origin by computing its centroid $\mathbf{\mu}_k^{(\mathcal{L})}$ and $d_\mathcal{L}(\mathbf{\mu}_k^{(\mathcal{L})}, \mathbf{o})$ using Eq.~\ref{eq:centroid}. We compute this distance for every data-point yielding a distribution $\mathcal{D}_{N_k}$. We then measure the overlap as the intersection area between different pair-wise kernel density estimates:
\begin{equation}
    \mathrm{OV}(N_k, N_{k'}) = \int \min\!\left(\hat{p}_{N_k}(x),\, 
    \hat{p}_{N_{k'}}(x)\right) dx
    \label{eq:ov}
\end{equation}
where $\hat{p}_{N}$ is the Gaussian kernel density estimate of the distance-to-origin values at level $N$. 
We repeat this operation for each pair, and obtain the final score by taking the mean.
$\mathrm{OV} \in [0, 1]$, where $0$ indicates perfectly disjoint distributions and $1$ indicates complete overlap. 

\noindent\textbf{Curvature reveals a separation--compression trade-off.}
Figures~\ref{fig:violin_slotcontrast}--\ref{fig:violin_spot} show the distribution of distances to the origin for three baselines across four manifolds: Euclidean, and Lorentz at $c \in \{0.2, 0.5, 1.0\}$. The most significant drop in inter-level overlap occurs at the first projection step from Euclidean to $c{=}0.2$, with OV reducing from $0.49$ to $0.41$ on SlotContrast, $0.62$ to $0.37$ on VideoSAURv2,  and $0.51$ to $0.36$ on SPOT. Increasing curvature to $c{=}0.5$ yields a further modest improvement across all three models (SlotContrast: $0.41{\to}0.38$, VideoSAUR-v2: $0.37{\to}0.36$, SPOT: $0.36{\to}0.35$), after which separation stabilises and OV values remain largely unchanged at $c{=}1.0$. Simultaneously, the violin shapes progressively collapse into narrow peaks, indicating that higher curvature compresses all representations toward the origin without further structuring them by level. This reveals a \emph{separation--compression tradeoff}: separation improves rapidly up to $c{=}0.5$ and then plateaus, while compression continues to increase with curvature. We therefore recommend $c{=}0.5$ as the practical default for post-hoc hyperbolic projection, as it captures the majority of the separation gain before the compression regime dominates.

\noindent\textbf{Consistent inverted depth ordering.}
Across all three baselines, coarse slots ($N{=}3$) lie furthest from the origin while fine slots ($N{=}13$) cluster closest. This is the opposite of what a supervised hyperbolic model would produce. We attribute this to the absence of hyperbolic supervision during training: Euclidean slot attention has no incentive to arrange representations by depth, and the post-hoc projection preserves this. The inversion has a plausible interpretation: coarse slots represent broad scene context, so their embeddings require more volume on the manifold; fine slots represent specific objects and cluster closer to the origin. End--to-end hyperbolic training could reverse this ordering.

\begin{figure}[hbt]
    \centering
    \vspace{-0.3cm}
    \includegraphics[width=0.45\textwidth]{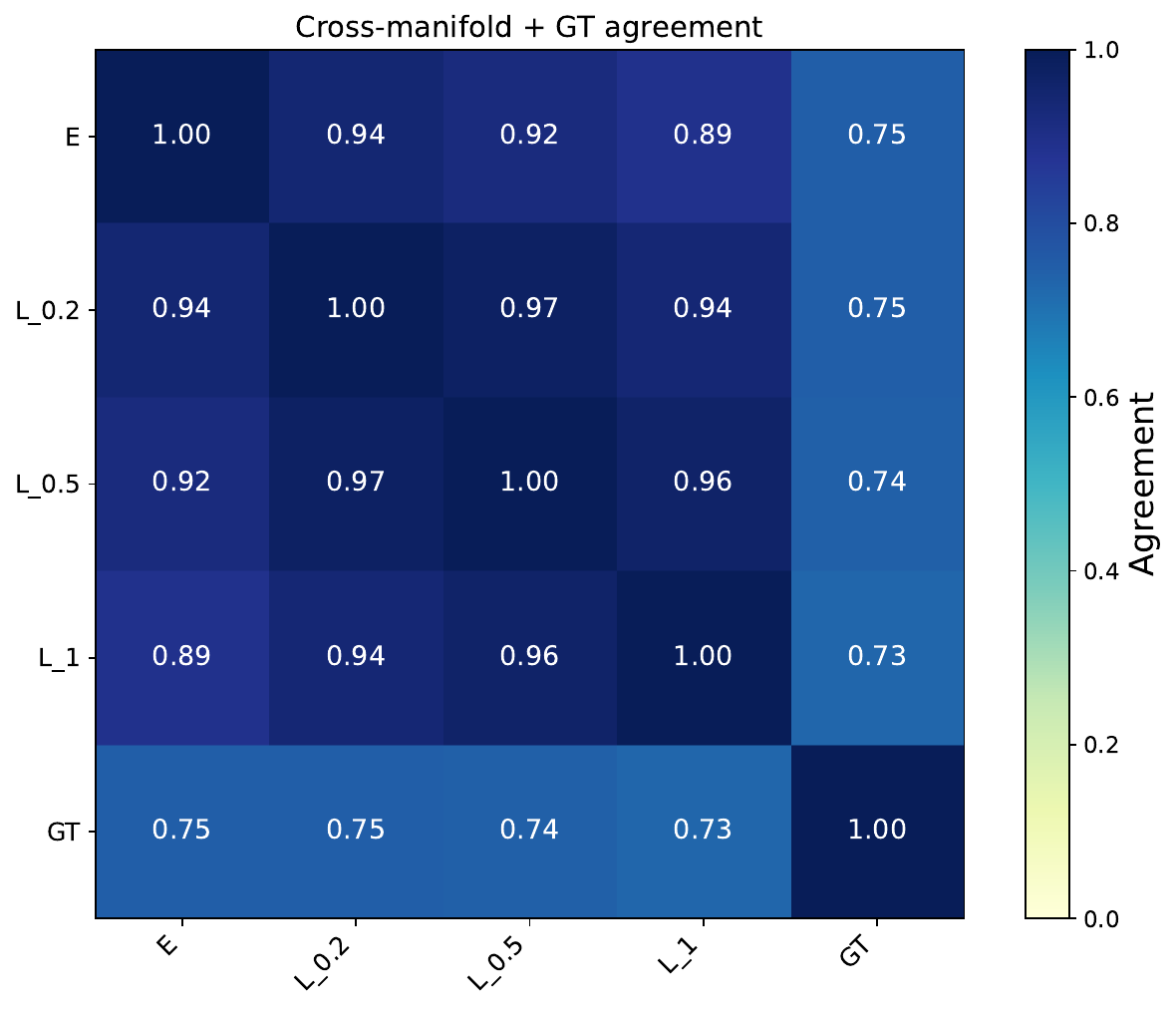}
    \vspace{-0.3cm}
    \caption{Pairwise agreement between hierarchical orderings induced by Euclidean, Lorentz at $c\in\{0.2, 0.5, 1.0\}$, and mask-based ground truth (GT) for SlotContrast. All manifolds agree substantially with GT (${\approx}0.75$) and Lorentz manifolds are mutually consistent ($0.89$--$0.94$), suggesting that both geometries capture the broad spatial organisation of the scene.}
    \vspace{-0.4cm}
    \label{fig:confusion}
\end{figure}

\noindent\textbf{Cross-manifold consistency.}
Figure~\ref{fig:confusion} reports pairwise agreement between the hierarchical orderings induced by each manifold and the mask-based ground truth, for Euclidean and Lorentz at $c \in \{0.2, 0.5, 1.0\}$ on SlotContrast (video). The three Lorentz manifolds are mutually consistent (agreement $0.89$--$0.94$), and all manifolds including Euclidean agree substantially with the mask-based GT hierarchy (all scores ${\approx}0.75$), indicating that both Euclidean and hyperbolic representations capture the broad spatial organisation of the scene. Differences between manifolds are subtle at this level of analysis; the more nuanced question of whether hyperbolic geometry provides a \emph{better} signal for individual parent--child associations is addressed directly via the parent retrieval task in Section~\ref{sec:retrieval}.

\subsection{Slot Norm Analysis}
\label{sec:norm_analysis}

\begin{table*}[ht]
    \centering
    \caption{Slot norms (mean$\,\pm\,$std) in Euclidean ($\ell_2$ norm 
    of raw slots) and Lorentz hyperbolic spaces ($c \in \{0.2, 0.5, 1\}$) 
    across granularity levels $N \in \{3, 5, 7, 11, 13\}$. Norms 
    \textbf{decrease monotonically with $N$} across all models, consistent 
    with the \textbf{inverted hierarchy}, observed in Figure~\ref{fig:violin}, which is exactly opposite of expected behaviour as presented in Figure \ref{fig:teaser}.}
    \vspace{-0.2cm}
    \label{tab:slot_norms}
    \setlength{\tabcolsep}{4pt}
    \renewcommand{\arraystretch}{1.2}
    \begin{tabular}{llccccc}
        \toprule
        \textbf{Model} & \textbf{Manifold}
            & $N=3$ & $N=5$ & $N=7$ & $N=11$ & $N=13$ \\
        \midrule
        \multirow{4}{*}{SlotContrast (video)}
        & $\|\mathbf{s}\|_2$
        & \std{1.446}{0.127}
        & \std{1.254}{0.137}
        & \std{1.193}{0.138}
        & \std{1.146}{0.140}
        & \std{1.137}{0.142} \\
        
        & $\mathbb{H}^d_{c=0.2}$
        & \std{1.089}{0.109}
        & \std{0.920}{0.105}
        & \std{0.858}{0.101}
        & \std{0.802}{0.100}
        & \std{0.789}{0.100} \\
        
        & $\mathbb{H}^d_{c=0.5}$
        & \std{0.817}{0.089}
        & \std{0.682}{0.079}
        & \std{0.630}{0.074}
        & \std{0.582}{0.073}
        & \std{0.570}{0.073} \\
        
        & $\mathbb{H}^d_{c=1}$
        & \std{0.609}{0.068}
        & \std{0.506}{0.059}
        & \std{0.465}{0.055}
        & \std{0.428}{0.055}
        & \std{0.418}{0.054} \\
        \midrule
        \multirow{4}{*}{VideoSAURv2(video)}
        & $\|\mathbf{s}\|_2$
        & \std{5.995}{0.192}
        & \std{5.835}{0.191}
        & \std{5.776}{0.195}
        & \std{5.741}{0.205}
        & \std{5.738}{0.205} \\

        & $\mathbb{H}^d_{c=0.2}$
        & \std{3.474}{0.220}
        & \std{3.100}{0.162}
        & \std{2.982}{0.146}
        & \std{2.906}{0.137}
        & \std{2.892}{0.139} \\

        & $\mathbb{H}^d_{c=0.5}$
        & \std{2.251}{0.153}
        & \std{1.994}{0.109}
        & \std{1.914}{0.096}
        & \std{1.860}{0.091}
        & \std{1.851}{0.093} \\

        & $\mathbb{H}^d_{c=1}$
        & \std{1.600}{0.108}
        & \std{1.415}{0.077}
        & \std{1.356}{0.068}
        & \std{1.317}{0.065}
        & \std{1.310}{0.068} \\
        \midrule
        \multirow{4}{*}{SPOT (image)}
        & $\|\mathbf{s}\|_2$
        & \std{5.669}{0.337}
        & \std{5.337}{0.332}
        & \std{5.191}{0.321}
        & \std{5.055}{0.323}
        & \std{5.016}{0.319} \\

        & $\mathbb{H}^d_{c=0.2}$
        & \std{2.455}{0.167}
        & \std{2.191}{0.144}
        & \std{2.087}{0.134}
        & \std{1.995}{0.129}
        & \std{1.970}{0.126} \\

        & $\mathbb{H}^d_{c=0.5}$
        & \std{1.574}{0.107}
        & \std{1.401}{0.092}
        & \std{1.333}{0.085}
        & \std{1.273}{0.082}
        & \std{1.257}{0.080} \\

        & $\mathbb{H}^d_{c=1}$
        & \std{1.124}{0.078}
        & \std{0.999}{0.066}
        & \std{0.949}{0.061}
        & \std{0.904}{0.059}
        & \std{0.893}{0.057} \\
        \bottomrule
    \end{tabular}
    \vspace{-0.2cm}
\end{table*}

Table~\ref{tab:slot_norms} reports mean slot norms across granularity levels $N \in \{3, 5, 7, 11, 13\}$ under Euclidean and Lorentz geometries for all models. Two consistent patterns emerge.

\noindent\textbf{Monotonic decrease with granularity.}
Both Euclidean and Lorentz norms decrease monotonically as $N$ increases across all models and curvature settings. This is consistent with the \emph{inverted depth ordering} observed in Section~\ref{sec:dilatation}: as the number of slots grows, each slot captures a smaller, more object-specific region of the scene and consequently lies closer to the origin on the hyperboloid --- the reverse of the root-near-origin ordering expected from supervised hyperbolic embeddings illustrated in Figure~\ref{fig:teaser}.
The decrease is smooth rather than abrupt, suggesting that slot representations gradually specialise with granularity rather than collapsing discontinuously.

\noindent\textbf{Hyperbolic projection amplifies level differences.}
While both geometries show the same monotonic trend, the relative spread between levels is more pronounced under Lorentz geometry. The ratio of norms between $N{=}3$ and $N{=}13$ is $1.45/1.14 \approx 1.23$ in Euclidean space for SlotContrast, versus $0.82/0.57 \approx 1.36$ at $c{=}0.5$. Similarly, on SPOT the Euclidean ratio is $5.67/5.02 \approx 1.13$, versus $1.57/1.26 \approx 1.25$ at $c{=}0.5$. This amplification effect is consistent across all models and grows with curvature: at $c{=}1$ the ratio increases further, though at the cost of compressing absolute norm values toward zero. This suggests that hyperbolic projection makes the geometric difference between coarse and fine slots more discriminative. Although the ordering itself is determined by the Euclidean representations, higher curvature shifts the emphasis from absolute scale to relative separability.


\begin{figure}[!b]
    \centering
    \vspace{-0.3cm}
    \includegraphics[width=0.48\textwidth]{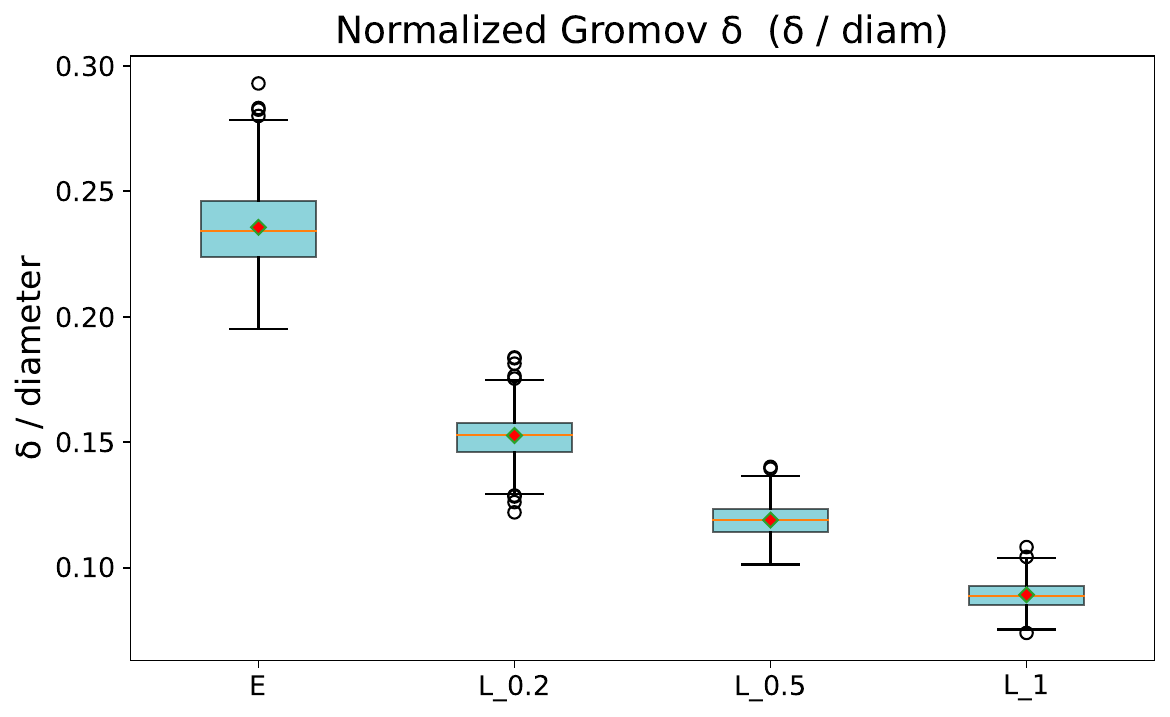}
    \vspace{-0.3cm}
    \caption{Distribution of normalized worst-case Gromov $\delta$ hyperbolicity from the slot embeddings of 300 videos of SlotContrast test set. A lower score (closer to 0) indicates that the pairwise distances between embedding encode tree-like structures. The progression shows how the score decreases as the curvature of the manifold increases.}
    \vspace{-0.4cm}
    \label{fig:gromov}
\end{figure}

\subsection{Normalized worst-case Gromov $\delta$ hyperbolicity}
Gromov $\delta$-hyperbolicity \cite{gromov} is a score that quantifies the tree-likeness of a set, and it has been widely used to measure hierarchies that emerge from graphs \cite{Tifrea2018PoincarGH}. 
In our case, these sets are slot-embeddings along with their pairwise distance matrix. 
We adopt a similar construction to the one proposed by Borassi \etal~\cite{PhysRevE.92.032812}, using the normalized worst-case Gromov $\delta$-hyperbolicity to measure hyperbolicity invariant to metric scaling, which is a property we need as the metric changes for each geometry. 
We repeat this operation for each sample in our dataset to obtain the hyperbolicity distributions shown in \ref{fig:gromov}. 
From the figure, we can observe that as the curvature increases, the hyperbolicity score of the embedded nodes approach $0$, suggesting that the structures that emerge from measuring the distances as curvature increases resemble more to \emph{tree-like hierarchies}.
This is contrary to the Euclidean case, where a higher score suggests that there are more interconnections between nodes, creating loops and breaking hierarchies. 
Although the slots were not trained to encode any hierarchy, these results suggests that the induced distance structure becomes increasingly tree-like. However, because spaces of higher negative curvature are inherently more hyperbolic, this trend may partially reflect properties of the target geometry rather than structure in the slot embeddings alone. We therefore interpret these results as preliminary evidence that hyperbolic projection can reveal hierarchical patterns in slot representations, rather than as definitive proof that the slots themselves encode such structure.


\subsection{Parent Slot Retrieval}
\label{sec:retrieval}
\vspace{-0.1cm}
\begin{table}[t]
    \centering
    \caption{Parent slot retrieval (Hit@1, \%) across consecutive level transitions, comparing Euclidean cosine distance and Lorentz geodesic at $c \in \{0.2, 0.5, 1.0\}$. Hit@1 measures the percentage of fine slots whose GT parent coarse slot is retrieved at rank 1. Best result per row in \textbf{bold}.}
    \vspace{-0.2cm}
    \label{tab:retrieval}
    \resizebox{0.98\columnwidth}{!}{%
    \setlength{\tabcolsep}{6pt}
    \renewcommand{\arraystretch}{1.2}
    \begin{tabular}{llcccc}
        \toprule
        & & \multicolumn{4}{c}{\textbf{Hit@1 (\%) $\uparrow$}} \\
        \cmidrule(lr){3-6}
        \textbf{Model} & \textbf{Manifold}
            & $3{\to}5$ & $5{\to}7$ & $7{\to}11$ & $11{\to}13$ \\
        \midrule
        \multirow{4}{*}{\rotatebox[origin=c]{70}{SlotContrast}}
            & Euclidean
            & 77.2 & 79.7          & \textbf{72.4} & \textbf{70.5} \\
            & $\mathbb{H}^d_{c=0.2}$
            & \textbf{77.8} & \textbf{79.9} & 71.8          & 70.2 \\
            & $\mathbb{H}^d_{c=0.5}$
            & 77.2          & 78.8          & 71.1          & 69.9 \\
            & $\mathbb{H}^d_{c=1.0}$
            & 76.1          & 77.4          & 69.3          & 68.7 \\
        \midrule
        \multirow{4}{*}{\rotatebox[origin=c]{70}{VideoSAURv2}}
            & Euclidean
            & 91.5          & 93.2          & \textbf{88.6} & \textbf{88.1} \\
            & $\mathbb{H}^d_{c=0.2}$
            & \textbf{91.9} & \textbf{93.3} & 88.5          & \textbf{88.1} \\
            & $\mathbb{H}^d_{c=0.5}$
            & 91.3          & 92.8          & 88.5          & \textbf{88.1} \\
            & $\mathbb{H}^d_{c=1.0}$
            & 90.8          & 92.4          & 87.8          & 88.0 \\
        \bottomrule
    \end{tabular}}
    \vspace{-0.4cm}
\end{table}

Given a fine-grained slot $\mathbf{s}_j^{(f)}$ at level $N_2$, we retrieve its \emph{predicted parent} from the set of coarse slots $\{\mathbf{s}_i^{(c)}\}_{i=1}^{N_1}$ by ranking all coarse slots by their distance to $\mathbf{s}_j^{(f)}$. Hit@1 measures the percentage of fine slots whose GT parent is the nearest coarse slot; a random baseline scores $1/N_1$ by definition (e.g.\ $33\%$ for the $3{\to}5$ transition where $N_1{=}3$), making observed scores above $70\%$ substantially above chance.

Table~\ref{tab:retrieval} shows that at low curvature ($c{=}0.2$), Lorentz geodesic distance matches or marginally outperforms Euclidean cosine on coarser transitions: Hit@1 improves from $79.7\%$ to $79.9\%$ on L5${\to}$L7 and from $77.2\%$ to $77.8\%$ on L3${\to}$L5 for SlotContrast, and from $93.2\%$ to $93.3\%$ on L5${\to}$L7 for VideoSAURv2, demonstrating that even a post-hoc hyperbolic projection introduces a useful geometric signal for parent--child association without any hyperbolic supervision. As curvature increases to $c{=}0.5$ and $c{=}1.0$, retrieval degrades progressively below Euclidean, as the depth penalty on the manifold increasingly dominates over angular similarity. 

\noindent\textbf{Curvature--task tradeoff.} Combined with the level separation results of Section~\ref{sec:dilatation}, where $c{=}0.5$ achieves the best inter-level disentanglement, this reveals a fundamental \emph{curvature--task tradeoff}: low curvature ($c{=}0.2$) is better for parent retrieval, preserving angular similarity while introducing a marginal depth signal; moderate curvature ($c{=}0.5$) is optimal for level separation, amplifying depth differences at the cost of retrieval precision. The two properties are actually complementary. We speculate that the norm separation reveals latent hierarchical structure that end-to-end hyperbolic training could fully exploit by using depth as an additional supervisory signal. Curvature is therefore a tunable parameter whose optimal value depends on the downstream application.

\vspace{-0.cm}
\subsection{Qualitative Analysis}
\label{sec:qualitative}
\vspace{-0.1cm}
\begin{figure*}[t]
    \vspace{-0.3cm}
    \centering
    \includegraphics[width=0.96\textwidth]{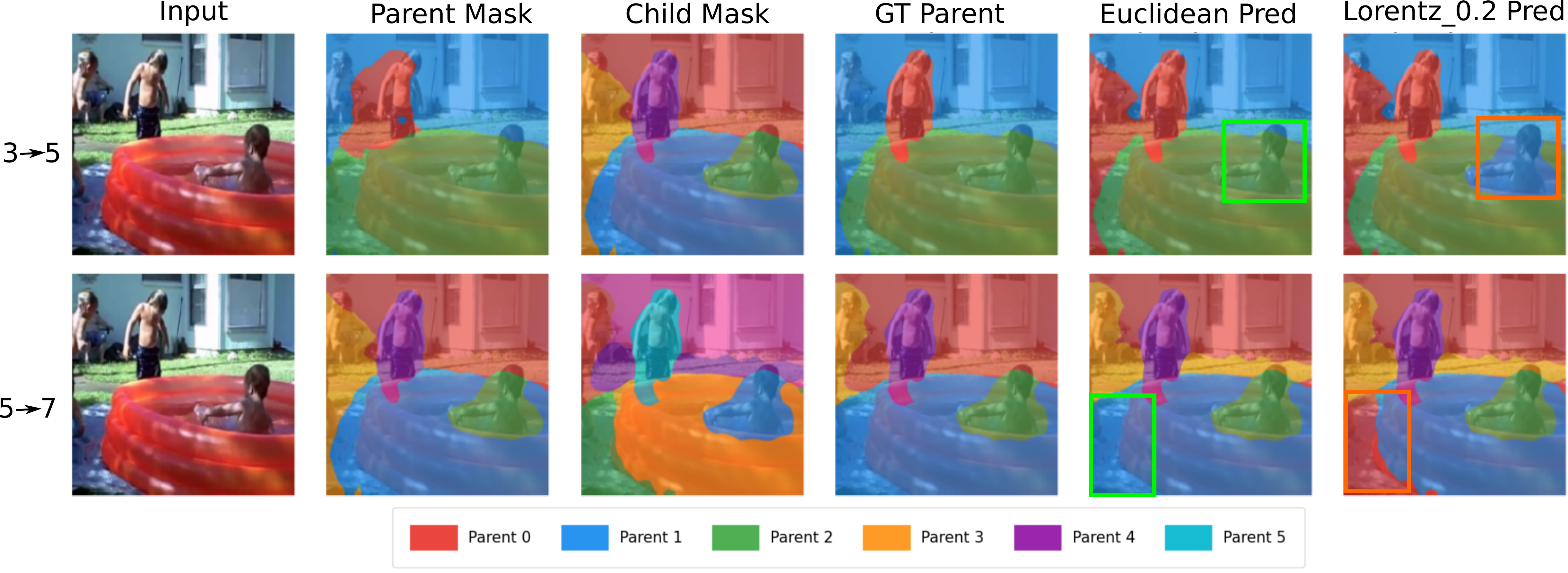}
    \vspace{-0.1cm}
    \caption{Parent slot retrieval for two-level transitions on a YTVIS video example for SlotContrast. Columns show the input, parent mask, child mask, GT assignment, Euclidean prediction, and Lorentz ($c{=}0.2$) prediction; correct predictions are highlighted in \green{green} and incorrect in \orange{orange}. Although Lorentz predictions are incorrect in some cases by the spatial GT criterion, they are semantically more meaningful, grouping slots by representational abstraction rather than spatial containment. Best viewed in colour.}
    \label{fig:qualitative}
    \vspace{-0.4cm}
\end{figure*}

Figure~\ref{fig:qualitative} highlights a key distinction between the two hierarchy notions considered in this work. The spatial ground-truth (GT) defines hierarchy through mask inclusion and spatial containment, whereas hyperbolic geometry organises slots according to representational abstraction in embedding space. The two approaches therefore agree in most cases (consistent with Table~\ref{tab:retrieval}), but diverge when spatial proximity and semantic relatedness do not align. Notably, several assignments counted as errors under the spatial GT correspond to semantically coherent groupings under the hyperbolic geometry.

The examples illustrate this behaviour. In the $3{\to}5$ transition, Euclidean cosine distance assigns a \emph{child slot} covering a child in the pool to the \emph{pool region}. This match is spatially plausible but semantically incorrect, driven by visual similarity between water and the partially submerged figure. Lorentz geodesic distance instead assigns the same \emph{child slot} to the \emph{background}. Although this is also incorrect under the spatial GT, it reflects a different geometric bias: hyperbolic projection favours broader contextual groupings over purely spatial adjacency.

A similar pattern appears in the $5{\to}7$ transition. The GT assigns a \emph{background slot} to the \emph{pool region} due to spatial proximity, which arises from the mask-inclusion criterion that conflates spatial containment with semantic relatedness. In contrast, the Lorentz geometry groups the \emph{background slot} with another \emph{background slot}, forming a semantically consistent background grouping.

Overall, these cases suggest that hierarchies capture complementary structural signals: spatial containment in the GT and representational abstraction in hyperbolic space.

\vspace{-0.1cm}
\section{Discussion}
\vspace{-0.2cm}
Our results show that projecting slot representations onto the Lorentz hyperboloid reveals a latent organizational structure that is hardly visible in Euclidean space. This structure is internally consistent across all curvature settings and models, suggesting it reflects a genuine representational property of the slots rather than a geometric artefact. Even without any hyperbolic supervision, low curvature ($c{=}0.2$) already matches or slightly improves parent retrieval over Euclidean, while higher curvatures achieve better level separation. This indicates that slot representations already carry latent hierarchical structure that hyperbolic geometry begins to expose. End-to-end hyperbolic training would provide the missing supervisory signal to fully align depth with hierarchy, turning this marginal improvement into a principled geometric encoding.

Furthermore, our analysis considers only adjacent level pairs for parent retrieval. A more complete picture would track each slot from leaf to root across all five granularity levels, enabling full scene decomposition trees that could directly improve the interpretability of object-centric models. Comparison with alternative hierarchy discovery methods such as clustering or tree reconstruction, and evaluation on downstream tasks, remain important future directions.
As slot representations were never trained with any hyperbolic objective, we observe an inverted depth ordering (from fine-to-coarse, instead of coarse-to-fine) and only marginal retrieval gains.

This work establishes a foundation for fully hyperbolic OCL. The natural next step is end-to-end training, where slot attention, positional encodings, and reconstruction are all performed in hyperbolic space. We expect this to correct the depth ordering and improve retrieval at higher curvatures. It would also enable evaluation beyond adjacent level pairs, including non-consecutive transitions and full leaf-to-root chains that build complete scene decomposition trees across all five granularity levels.
This will then provide a principled geometric framework for hierarchical visual understanding. Such structured slot hierarchies could support more interpretable and temporally consistent video object and instance segmentation, as well as improve semantic and panoptic segmentation in videos by modeling part–whole relationships across frames. More broadly, hyperbolic representations may offer a scalable way to capture multi-level scene structure, with direct relevance to autonomous driving, robotic perception, and video understanding in complex environments.

\vspace{-0.1cm}
\section{Conclusion}
\vspace{-0.1cm}
We showed that projecting Euclidean slot representations onto the Lorentz hyperboloid reveals latent hierarchical structure in object-centric scene representations. Applying our pipeline to three baselines across image and video domains, we consistently find that hyperbolic geometry exposes a coarse-to-fine organisation among slots that are hardly visible in Euclidean space, with coarse slots occupying greater manifold depth than fine slots. We identify a curvature--task tradeoff: low curvature ($c{=}0.2$) matches or outperforms Euclidean on the parent slot retrieval task, while moderate curvature ($c{=}0.5$) achieves the best inter-level separation. Beyond retrieval performance, our results reveal a qualitative difference in how the two geometries construct hierarchies: Euclidean representations emphasize \emph{spatial proximity}, while hyperbolic representations capture \emph{semantic relatedness}.
This gap between spatial and semantic hierarchy motivates end-to-end hyperbolic object-centric learning as a principled next step toward fully geometric hierarchical scene understanding.

\vspace{-0.1cm}
\section*{Acknowledgements}
\vspace{-0.2cm}
The authors thank Rongzhen Zhao for discussions on hierarchical object-centric learning. This work was supported by the Innovation Fund Denmark, Grant No. 2081-00001B; MRPA; the Spanish project PID2022-136436NB-I00; ICREA under the ICREA Academia programme; and the Canada CIFAR AI Chairs.

{
    \small
    \bibliographystyle{ieeenat_fullname}
    \bibliography{main}
}

\end{document}